\documentclass{article}

\usepackage{graphicx}
\usepackage{amsmath}

\title{Discovering network behind infectious disease outbreak}

\author{Yoshiharu Maeno \\ Social Design Group \\ email: maeno.yoshiharu@socialdesigngroup.com}

\begin{document}

\maketitle

\begin{abstract}
Stochasticity and spatial heterogeneity are of great interest recently in studying the spread of an infectious disease. The presented method solves an inverse problem to discover the effectively decisive topology of a heterogeneous network and reveal the transmission parameters which govern the stochastic spreads over the network from a dataset on an infectious disease outbreak in the early growth phase. Populations in a combination of epidemiological compartment models and a meta-population network model are described by stochastic differential equations. Probability density functions are derived from the equations and used for the maximal likelihood estimation of the topology and parameters. The method is tested with computationally synthesized datasets and the WHO dataset on SARS outbreak.
\end{abstract}

\section{Introduction}
\label{problem}

When the epidemiologists at a public health agency detect a signal of an infectious disease outbreak, they rely heavily on mathematical models of disease transmission in estimating the rate of transmission, predicting the direction and speed of the spread, and figuring out an effective measure to contain the outbreak. Many of the models formulate stochasticity and spatial heterogeneity, which are of great interest recently. The spatial heterogeneity ranges from the uneven probabilities of contacts between the individuals in communities \cite{Wal10}, \cite{Sma06}, dependence of the strength of the demographical interactions between cities on the distance \cite{Kee04}, to nation-wide or world-wide inhomogeneous geographical structures \cite{Dan09}, \cite{Ril07}.

A Monte-Carlo stochastic simulation is widely used to understand the influence of the spatial heterogeneity on stochastic spreading. In such a simulation, accuracy and reproducibility of the input demographical knowledge such as the amount of traffic between cities have great impacts on the reliability of the output pattern of the movement of pathogens and their hosts. {\em But, in studying world-wide epidemics, just a collection of regular airline routes and aircraft capacities does not always present the transportation network which results in the real chain of transmission. Some routes are influential decisively, but the others are not. Examples are found in the spread of Severe Acute Respiratory Syndrome (SARS) from Asia to the world in 2003. There were not any cases in Japan in spite of the heavy traffic there from Asian countries. Many patients appeared in Canada earlier than the United States to which airlines connect Asian countries much more densely. Here arises an interesting question. Inversely, is it possible to learn the effectively decisive transportation network by observing how the disease spreads, reinforce the demographical knowledge on the network, and import the acquired knowledge into the mathematical model?} This is an inverse problem similar to the network tomography \cite{Mae09}, \cite{Rab08}.

In this study, a statistical method is presented to discover the effectively decisive topology of a heterogeneous network and reveal the parameters which govern stochastic transmission from a dataset on the early growth phase of the outbreak. The dataset consists of either the number of infectious persons or the number of new cases per an observation interval. The method is founded on a mathematical model for a stochastic reaction-diffusion process \cite{Bar08}. The population in the model is described by a set of Langevin equations. The equations are stochastic differential equations which include rapidly fluctuating and highly irregular functions of time. Probability density functions and likelihood functions are derived from the equations analytically, and used for the maximal likelihood estimation of the topology and parameters. The method is tested with a number of computationally synthesized datasets and the World Health Organization (WHO) dataset on the SARS outbreak in March through April in 2003.

\section{Problem}
\label{problemDefinition}

\subsection{Stochastic model}
\label{spread}

The model in this study is a special case of a stochastic reaction-diffusion process. The model is a combination of standard epidemiological SIR or SIS compartment models and a meta-population network model. The meta-population network model \cite{Col07} sub-divides the entire population into distinct sub-populations in $N$ geographical regions. Movement of persons occurs between the sub-populations while the epidemiological state transitions (infection and recovery) occur in a sub-population. A sub-population is randomly well-mixed. Heterogeneity is present between sub-populations.

The geographical regions are represented by nodes $n_{i}\ (i=0, 1, \cdots, N-1)$. {\em The movement is parameterized by a matrix $\mbox{\boldmath{$\gamma$}}$ whose $i$-th row and $j$-th column element $\gamma_{ij}$ is the probability at which a person moves from $n_{i}$ to $n_{j}$ per a unit time. A person remains at the same node at the probability of $1-\sum_{j=0}^{N-1} \gamma_{ij}$. Generally, $\gamma_{ij}=\gamma_{ji}$ does not hold. By definition, $\gamma_{ii}=0$. It is often confirmed empirically that a simple law relates a network topology to the movement \cite{Bar04}. The topology is specified by a neighbor matrix $\mbox{\boldmath{$l$}}$. The transportation between two regions is represented by a pair of unidirectional links. If a pair of links is present between $n_{i}$ and $n_{j}$, $l_{ij}=l_{ji}=1$. If absent, $l_{ij}=l_{ji}=0$. By definition, $l_{ii}=0$. In the experiments in section \ref{experiment}, an empirically confirmed law $\gamma_{ij}=\Gamma_{ij}(\mbox{\boldmath{$l$}})$ is postulated, and the topology and the probability of movement are treated interchangeably.}

The SIR compartment model \cite{Kee08} is a behavioral extreme where immunity is life-long. The state of a person changes from a susceptible state ($S$), through an infectious state ($I$), to a recovered state ($R$). In contrast, the immunity does not occur in the SIS compartment model. The state of a recovered patient goes back to $S$. The parameter $\alpha$ represents the probability at which an infectious person contacts a person and infect the person per a unit time. If the contacted person is susceptible, the number of the infectious persons increases by 1. The effective rate of infection by a single infectious person is the product of $\alpha$ and the proportion of the susceptible persons within the population. The parameter $\beta$ represents the probability at which an infectious person recovers per a unit time. These parameters are constants over subpopulations and time. The basic reproductive ratio $r$ is defined by $r=\alpha/\beta$ \cite{Lip03}.

Movement, infection and recovery are Markovian stochastic processes governed by $\gamma_{ij}$, $\alpha$, and $\beta$.

\subsection{Time evolution of spread}

In a stochastic process, even if the initial condition is known, there are many possible trajectories which the process might go along. A set of these possible trajectories is a statistical ensemble. The change in the population is described by a set of Langevin equations \cite{Huf04}. A Langevin equation is a stochastic differential equation \cite{Klo92}. The microscopic continuous time evolution of a system is obtained by adding a fluctuation (a stochastic term) to the known macroscopic time evolution of the system.

The quantity $S_{i}(t)$ is the number of susceptible persons at a node $n_{i}$ at time $t$. $I_{i}(t)$ is the number of infectious persons. $R_{i}(t)$ is the number of recovered persons. The change in $I_{i}(t) \ (i=0,1, \cdots, N-1)$ is given by eq.(\ref{dI/dt}) \cite{Col06}. It is a set of $N$ stochastic differential equations.
\begin{eqnarray}
\frac{{\rm d} I_{i}(t)}{{\rm d} t} &=& \frac{\alpha S_{i}(t) I_{i}(t)}{S_{i}(t)+I_{i}(t)+R_{i}(t)} - \beta I_{i}(t) 
+ \sum_{j=0}^{N-1} \gamma_{ji} I_{j}(t) - \sum_{j=0}^{N-1} \gamma_{ij} I_{i}(t) \nonumber \\
&+& \sqrt{\frac{\alpha S_{i}(t) I_{i}(t)}{S_{i}(t)+I_{i}(t)+R_{i}(t)}} \xi^{[\alpha]}_{i}(t) 
- \sqrt{\beta I_{i}(t)} \xi^{[\beta]}_{i}(t) \nonumber \\
&+& \sum_{j=0}^{N-1} \sqrt{\gamma_{ji} I_{j}(t)} \xi^{[\gamma]}_{ji}(t) - \sum_{j=0}^{N-1} \sqrt{\gamma_{ij} I_{i}(t)} \xi^{[\gamma]}_{ij}(t).
\label{dI/dt}
\end{eqnarray}

Stochastic terms $\mbox{\boldmath{$\xi$}}(t) = ( \xi^{[\alpha]}_{i}(t), \xi^{[\beta]}_{i}(t), \xi^{[\gamma]}_{ij}(t) )$ are rapidly fluctuating and highly irregular functions of time. {\em The number of terms is $M=N^{2}+N$ ($N$ terms for infection, $N$ terms for recovery, and $N(N-1)$ terms for movement). The functional forms of individual elements $\xi_{a}(t) \ (a=0,1, \cdots, M-1)$ are not known. Their statistical property is the Gaussian white noise which satisfies eq.(\ref{gWhite2}) through (\ref{gWhite4}).}
\begin{eqnarray}
\langle \xi_{a}(t) \rangle_{{\rm ensemble}} =0.
\label{gWhite2}
\end{eqnarray}
\begin{eqnarray}
\langle \xi_{a}(t) \xi_{b}(u) \rangle_{{\rm ensemble}} = \delta_{ab} \delta(u-t).
\label{gWhite3}
\end{eqnarray}
\begin{eqnarray}
\langle \xi_{a}(t) \xi_{b}(u) \xi_{c}(v) \cdots \rangle_{{\rm ensemble}} = 0.
\label{gWhite4}
\end{eqnarray}

In these equations, $\delta(t)$ is a Dirac's delta function, and $\delta_{ab}$ is a Kronecker's delta symbol. The ensemble average of a variable $x$ is $\langle x \rangle_{{\rm ensemble}}$. Eq.(\ref{gWhite3}) means that there is no correlation at different times and between different terms. Eq.(\ref{gWhite4}) means that the third and higher order moments vanish.

In most cases, the outbreak is contained before the spread reaches equilibrium. In the early growth phase of the outbreak, $I_{i} \ll S_{i}$ and $R_{i} \ll S_{i}$ hold true. The first term of the rightside of eq.(\ref{dI/dt}) is independent of $S_{i}$ and $R_{i}$ because $S_{i}/(S_{i}+I_{i}+R_{i}) \approx 1$. The resulting equation is eq.(\ref{apdI/dt}). Eq.(\ref{apdI/dt}) can also be applied to the SIS model.
\begin{eqnarray}
\frac{{\rm d} I_{i}(t)}{{\rm d} t} &=& \alpha I_{i}(t) - \beta I_{i}(t) 
+ \sum_{j=0}^{N-1} \gamma_{ji} I_{j}(t) - \sum_{j=0}^{N-1} \gamma_{ij} I_{i}(t) \nonumber \\
&+& 
\sqrt{\alpha I_{i}(t)} \xi^{[\alpha]}_{i}(t) - \sqrt{\beta I_{i}(t)} \xi^{[\beta]}_{i}(t) \nonumber \\
&+& \sum_{j=0}^{N-1} \sqrt{\gamma_{ji} I_{j}(t)} \xi^{[\gamma]}_{ji}(t) - \sum_{j=0}^{N-1} \sqrt{\gamma_{ij} I_{i}(t)} \xi^{[\gamma]}_{ij}(t).
\label{apdI/dt}
\end{eqnarray}

The cumulative number of new cases until time $t$ is represented by $J_{i}(t) \ (i=0, 1, \cdots, N-1)$. The rate of increase in $J_{i}(t)$ equals to the first term of eq.(\ref{apdI/dt}). That is, $\alpha I_{i}(t)$. The time evolution of $J_{i}(t)$ is given by eq.(\ref{dJ/dt}). The rightside dose not depend on $J_{i}(t)$ itself.
\begin{eqnarray}
\frac{{\rm d} J_{i}(t)}{{\rm d} t} = \alpha I_{i}(t) + \sqrt{\alpha I_{i}(t)} \xi^{[\alpha]}_{i}(t).
\label{dJ/dt}
\end{eqnarray}

The total number of the infectious persons at time $t$ is given by $I(t)=\sum_{i=0}^{N-1} I_{i}(t)$. Its time evolution is given by eq.(\ref{dIall/dt}). It does not depend on the values of $\gamma_{ij}$.
\begin{eqnarray}
\frac{{\rm d} I(t)}{{\rm d} t} = \alpha I(t) - \beta I(t) + \sum_{i=0}^{N-1} \sqrt{\alpha I_{i}(t)} \xi^{[\alpha]}_{i}(t) - \sum_{i=0}^{N-1} \sqrt{\beta I_{i}(t)} \xi^{[\beta]}_{i}(t).
\label{dIall/dt}
\end{eqnarray}

The total cumulative number of new cases until time $t$ is given by $J(t)=\sum_{i=0}^{N-1} J_{i}(t)$. Its time evolution is given by eq.(\ref{dJall/dt}). 
\begin{eqnarray}
\frac{{\rm d} J(t)}{{\rm d} t} = \alpha I(t) + \sum_{i=0}^{N-1} \sqrt{\alpha I_{i}(t)} \xi^{[\alpha]}_{i}(t).
\label{dJall/dt}
\end{eqnarray}

\subsection{Definition of problem}
\label{profiling}

The problem is to discover the network topology $\mbox{\boldmath{$l$}}$ (or $\mbox{\boldmath{$\gamma$}}$) and reveal the transmission parameter $r$ (or $\alpha$ and $\beta$) from a given dataset $I_{i}(t_{d}) \ (i=0,1,\cdots,N-1, \ d=0,1,\cdots,D-1)$ or $\Delta J_{i}(t_{d}) \ (i=0,1,\cdots,N-1, \ d=0,1,\cdots,D-1)$. The dataset $I_{i}(t_{d})$ is the time sequence of the number of infectious persons. The dataset $\Delta J_{i}(t_{d}) = J_{i}(t_{d+1}) - J_{i}(t_{d})$ is the time sequence of the number of new cases between observations. Observation is made at every node $n_{i}\ (i=0,1,\cdots,N-1)$ at times $t_{d} \ (d=0,1,\cdots,D-1)$. The time interval between observations is $\Delta t=t_{d+1}-t_{d}$. For example, a bundle of the daily reports on cases from hospitals is a dataset $\Delta J_{i}(t_{d})$ where $\Delta t=1$ day. Other information is not known. That is, nothing is known about $S_{i}(t_{d})$, $R_{i}(t_{d})$, nor the initial condition which could identify the index case (the first patient from whom the infectious disease has spread). 

\section{Method}
\label{method}

\subsection{Likelihood function}
\label{pdf}

Various techniques of statistical inference can be applied once the likelihood function is obtained analytically. {\em The likelihood function is the conditional probability of the obtained dataset as a function of the unknown parameters of a parameterized statistical model. The conditional probability becomes noticeably large if the value of the parameters is close to the true value.} For example, maximal a posteriori estimation is used to find the parameters which maximize the posterior distribution. In this study, the problem is solved by maximal likelihood estimation. The Langevin equations (\ref{apdI/dt}) through (\ref{dJall/dt}) are solved by obtaining the moments of probability variables at time $t$ so that the probability density functions and logarithmic likelihood functions can be derived, rather than by calculating the trajectories of time-dependent variables for a given functional form of $\xi_{a}(t)$ \cite{Dan09}. Four logarithmic likelihood functions $L^{{\rm [I1]}}(\mbox{\boldmath{$\theta$}})$, $L^{{\rm [I2]}}(\mbox{\boldmath{$\theta$}})$, $L^{{\rm [J1]}}(\mbox{\boldmath{$\theta$}})$, and $L^{{\rm [J2]}}(\mbox{\boldmath{$\theta$}})$ are derived for given datasets $I_{i}(t_{d})$, $I(t_{d})$, $\Delta J_{i}(t_{d})$, and $\Delta J(t_{d})$ respectively under the unknown parameters $\mbox{\boldmath{$\theta$}}=\{\mbox{\boldmath{$\gamma$}}, \alpha, \beta \}$.

Appendix \ref{Solving} presents the procedure to solve the Langevin equations through a Fokker-Planck equation \cite{Kam07} . Appendix \ref{moIiJi} summarizes the formula for the time evolution of $\mbox{\boldmath{$m$}}^{{\rm [I]}}(t|\mbox{\boldmath{$\theta$}})$, $\mbox{\boldmath{$m$}}^{{\rm [J]}}(t|\mbox{\boldmath{$\theta$}})$ (row vectors whose $i$-th element is the mean of $I_{i}$, $J_{i}$) and $\mbox{\boldmath{$v$}}^{{\rm [II]}}(t|\mbox{\boldmath{$\theta$}})$, $\mbox{\boldmath{$v$}}^{{\rm [IJ]}}(t|\mbox{\boldmath{$\theta$}})$, $\mbox{\boldmath{$v$}}^{{\rm [JJ]}}(t|\mbox{\boldmath{$\theta$}})$ (matrices whose $i$-th row and $j$-th column element is the covariance between $I_{i}$ and $I_{j}$, $I_{i}$ and $J_{j}$, $J_{i}$ and $J_{j}$). Appendix \ref{moIJ} summarizes the formula for the time evolution of $m^{{\rm [I]}}(t|\mbox{\boldmath{$\theta$}})$, $m^{{\rm [J]}}(t|\mbox{\boldmath{$\theta$}})$ (the mean of $I$, $J$) and $v^{{\rm [II]}}(t|\mbox{\boldmath{$\theta$}})$, $v^{{\rm [IJ]}}(t|\mbox{\boldmath{$\theta$}})$, $v^{{\rm [JJ]}}(t|\mbox{\boldmath{$\theta$}})$ (the variance of $I$, covariance between $I$ and $J$, variance of $J$).

\subsubsection{Case 1: for given $I_{i}(t_{d})$}
\label{pdffromI}

The logarithmic likelihood function $L^{[{\rm I1}]}(\mbox{\boldmath{$\theta$}})$ is determined by $I_{i}(t_{d})$. If the third and higher order moments are ignored, the probability density function $p(\mbox{\boldmath{$I$}},t_{d+1}|\mbox{\boldmath{$\theta$}})$ is a multi-variate Gaussian distribution with the mean $\mbox{\boldmath{$m$}}^{{\rm [I]}}(t_{d+1}|\mbox{\boldmath{$\theta$}})$ and covariance $\mbox{\boldmath{$v$}}^{{\rm [II]}}(t_{d+1}|\mbox{\boldmath{$\theta$}})$ in eq.(\ref{probsolI}). This is the probability of $\mbox{\boldmath{$I$}} = (I_{0},\cdots,I_{n-1})$ at $t=t_{d+1}$ given $\mbox{\boldmath{$\theta$}}$. $\mbox{\boldmath{$X$}}^{T}$ is a transpose of a matrix \mbox{\boldmath{$X$}}.
\begin{eqnarray}
p^{{\rm [I1]}}(\mbox{\boldmath{$I$}},t_{d+1}|\mbox{\boldmath{$\theta$}}) = \frac{ \exp (-\frac{1}{2} (\mbox{\boldmath{$I$}}-\mbox{\boldmath{$m$}}^{{\rm [I]}}(t_{d+1}|\mbox{\boldmath{$\theta$}})) \mbox{\boldmath{$v$}}^{{\rm [II]}}(t_{d+1}|\mbox{\boldmath{$\theta$}})^{-1} (\mbox{\boldmath{$I$}}-\mbox{\boldmath{$m$}}^{{\rm [I]}}(t_{d+1}|\mbox{\boldmath{$\theta$}}))^{T}) }{\sqrt{(2\pi)^{N} \det \mbox{\boldmath{$v$}}^{{\rm [II]}}(t_{d+1}|\mbox{\boldmath{$\theta$}}) }}.
\label{probsolI}
\end{eqnarray}

$L^{[{\rm I1}]}(\mbox{\boldmath{$\theta$}})$ is the logarithm of a product of the probability of the individual observation $\mbox{\boldmath{$I$}}(t_{d+1})$ at $t=t_{d+1}$. It is given by eq.(\ref{L[I1]}).
\begin{eqnarray}
L^{[{\rm I1}]}(\mbox{\boldmath{$\theta$}}) = \sum_{d=0}^{D-2} \log p^{{\rm [I1]}}(\mbox{\boldmath{$I$}}(t_{d+1}),t_{d+1}|\mbox{\boldmath{$\theta$}}). 
\label{L[I1]}
\end{eqnarray}

If $\Delta t$ is small, the formula for the moments become simpler. The exact formula for $\mbox{\boldmath{$m$}}^{{\rm [I]}}(t|\mbox{\boldmath{$\theta$}})$ and $\mbox{\boldmath{$v$}}^{{\rm [II]}}(t|\mbox{\boldmath{$\theta$}})$ are expanded in terms of $\Delta t$, and the second and higher order terms are ignored. If the data $I_{i}(t_{d})$ at $t=t_{d}$ is reliable completely, $m^{{\rm [I]}}_{i}(t_{d}|\mbox{\boldmath{$\theta$}})=I_{i}(t_{d})$ and $v^{{\rm [II]}}_{ij}(t_{d}|\mbox{\boldmath{$\theta$}})=0$. The moments after $\Delta t$ are given by eq.(\ref{apmIisol}) and (\ref{apvIIijsol}). The moments at $t=t_{d+1}$ depend only on the data at $t=t_{d}$.
\begin{eqnarray}
m^{{\rm [I]}}_{i}(t_{d+1}|\mbox{\boldmath{$\theta$}}) \approx I_{i}(t_{d}) 
+ \{ (\alpha-\beta-\sum_{k=0}^{N-1} \gamma_{ik}) I_{i}(t_{d}) + \sum_{j=0}^{N-1} \gamma_{ji} I_{j}(t_{d}) \} \Delta t.
\label{apmIisol}
\end{eqnarray}
\begin{eqnarray}
v^{{\rm [II]}}_{ij}(t_{d+1}|\mbox{\boldmath{$\theta$}}) &\approx& 
[ \{(\alpha+\beta+\sum_{k=0}^{N-1} \gamma_{ik}) I_{i}(t_{d}) +\sum_{k=0}^{N-1} \gamma_{ki} I_{k}(t_{d}) \} \delta_{ij} \nonumber \\
&-& \gamma_{ij} I_{i}(t_{d}) - \gamma_{ji} I_{j}(t_{d}) ] \Delta t.
\label{apvIIijsol}
\end{eqnarray}

Similarly, the logarithmic likelihood function $L^{{\rm [I2]}}(\mbox{\boldmath{$\theta$}})$ is determined by $I(t_{d})$. $I(t_{d})$ can be calculated from the given dataset $I_{i}(t_{d})$. The probability density function $p^{{\rm [I2]}}(I,t_{d+1}|\mbox{\boldmath{$\theta$}})=p^{{\rm [I2]}}(I,t_{d+1}|\alpha,\beta)$ is a Gaussian distribution with the mean $m^{{\rm [I]}}(t_{d+1}|\mbox{\boldmath{$\theta$}})$ and variance $v^{{\rm [II]}}(t_{d+1}|\mbox{\boldmath{$\theta$}})$. It does not depend on $\mbox{\boldmath{$\gamma$}}$. $L^{{\rm [I2]}}(\mbox{\boldmath{$\theta$}})=L^{[{\rm I2}]}(\alpha,\beta)$ is the logarithm of a product of the probability of individual observation $I(t_{d+1})$ at $t=t_{d+1}$. It is given by eq.(\ref{L[I2]}). 
\begin{eqnarray}
L^{[{\rm I2}]}(\alpha,\beta) = \sum_{d=0}^{D-2} \log p^{{\rm [I2]}}(I(t_{d+1}),t_{d+1}|\alpha,\beta). 
\label{L[I2]}
\end{eqnarray}

Again, if $\Delta t$ is small, the formula for the moments become simpler. They are given by eq.(\ref{apmIsol}) and (\ref{apvIIsol}).
\begin{eqnarray}
m^{{\rm [I]}}(t_{d+1}|\mbox{\boldmath{$\theta$}}) \approx I(t_{d}) + (\alpha-\beta) I(t_{d}) \Delta t.
\label{apmIsol}
\end{eqnarray}
\begin{eqnarray}
v^{{\rm [II]}}(t_{d+1}|\mbox{\boldmath{$\theta$}}) \approx (\alpha+\beta) I(t_{d}) \Delta t.
\label{apvIIsol}
\end{eqnarray}

\subsubsection{Case 2: for given $\Delta J_{i}(t_{d})$}
\label{pdffromJ}

The logarithmic likelihood function $L^{[{\rm J1}]}(\mbox{\boldmath{$\theta$}})$ is determined by $\Delta J_{i}(t_{d})$. The probability density function $p^{{\rm [J1]}}(\mbox{\boldmath{$J$}},t_{d+1}|\mbox{\boldmath{$\theta$}})$ is a multi-variate Gaussian distribution in the same functional form as eq.(\ref{probsolI}) with the mean $\mbox{\boldmath{$m$}}^{{\rm [J]}}(t_{d+1}|\mbox{\boldmath{$\theta$}})$ and covariance $\mbox{\boldmath{$v$}}^{{\rm [JJ]}}(t_{d+1}|\mbox{\boldmath{$\theta$}})$. The functional form of $L^{[{\rm J1}]}$ is also the same as eq.(\ref{L[I1]}). $J_{i}(t_{d+1})$ can be calculated from the dataset by $J_{i}(t_{d+1})=J_{i}(t_{0})+\sum_{d'=0}^{d} \Delta J_{i}(t_{d'})$. The first observation $\Delta J_{i}(t_{0})$ may include all the known cases at that time. Then, $J_{i}(t_{0})$ can be deleted from the above formula. There is, however, a big difference from $L^{[{\rm I1}]}(\mbox{\boldmath{$\theta$}})$. The probability at $t=t_{d+1}$ is not determined by the data $J_{i}(t_{d})$ because the moments of $J_{i}$ at $t=t_{d+1}$ depends on $I_{i}$ at $t=t_{d}$ whose value is not known. Such approximation as eq.(\ref{apmIsol}) or (\ref{apvIIsol}) is not correct. Thus, the exact formula for $\mbox{\boldmath{$v$}}^{{\rm [JJ]}}(t|\mbox{\boldmath{$\theta$}})$ as a function of $t$ must be evaluated to calculate the value of $L^{[{\rm J1}]}(\mbox{\boldmath{$\theta$}})$.

Similarly, the logarithmic likelihood function $L^{[{\rm J2}]}(\mbox{\boldmath{$\theta$}})$ is determined by $J(t_{d})$. $J(t_{d})$ can be calculated from the given dataset $\Delta J_{i}(t_{d})$. The probability density function $p^{{\rm [J2]}}(J,t_{d+1}|\mbox{\boldmath{$\theta$}})=p^{{\rm [J2]}}(J,t_{d+1}|\alpha,\beta)$ is a Gaussian distribution with the mean $m^{{\rm [J]}}(t_{d+1}|\mbox{\boldmath{$\theta$}})$ and variance $v^{{\rm [JJ]}}(t_{d+1}|\mbox{\boldmath{$\theta$}})$. The functional form of $L^{[{\rm J2}]}(\mbox{\boldmath{$\theta$}})=L^{[{\rm J2}]}(\alpha,\beta)$ is the same as eq.(\ref{L[I2]}). The approximation in eq.(\ref{apmIsol}) and (\ref{apvIIsol}) is not correct either. Thus, the exact formula for $v^{{\rm [JJ]}}(t|\mbox{\boldmath{$\theta$}})$ as a function of $t$ must be evaluated to calculate the value of $L^{[{\rm J2}]}(\alpha,\beta)$.

\subsection{Estimation procedure}
\label{estimation}

{\em Theoretically, every formula in the following can be applied in estimating $\gamma_{ij}$ directly, as well as estimating $\mbox{\boldmath{$l$}}$ and obtaining $\mbox{\boldmath{$\gamma$}}$ by the law $\gamma_{ij}=\Gamma_{ij}(\mbox{\boldmath{$l$}})$. But, the estimation of $N(N-1)/2$ binary parameters $l_{ij} \ (i < j)$ tends to be more robust than that of $N(N-1)$ continuous parameters $\gamma_{ij} \ (i \neq j)$. The binary parameters are suitable for reliable combinatorial optimization by means of well-established numerical algorithms and computational implementations. Thus, the estimation of $\mbox{\boldmath{$l$}}$ is detailed here and demonstrated in section \ref{experiment}.}

\subsubsection{Case 1: for given $I_{i}(t_{d})$}
\label{estimationfromI}

The procedure for the estimation from $I_{i}(t_{d})$ is prsented. The problem is solved by dividing it to two sub-problems and solving them sequentially, rather than by searching the maximal likelihood estimators $\hat{\alpha}$, $\hat{\beta}$, and $\hat{\mbox{\boldmath{$l$}}}$ simultaneously. The first sub-problem is to obtain $\hat{\alpha}$ and $\hat{\beta}$ by solving eq.(\ref{esalbe}).
\begin{eqnarray}
\hat{\alpha}, \ \hat{\beta} = \arg \underset{\alpha,\ \beta}{\max} L^{[{\rm I2}]}(\alpha,\beta).
\label{esalbe}
\end{eqnarray}

The estimators are given by eq.(\ref{alphaestimator}) and (\ref{betaestimator}) where $\Delta I(t_{d}) = I(t_{d+1})-I(t_{d})$.
\begin{eqnarray}
\hat{\alpha} = \frac{1}{2 \Delta t} \{ \frac{1}{D} \sum_{d=0}^{D-1}  \frac{\Delta I(t_{d})}{I(t_{d})} - \frac{1}{D} \frac{(\sum_{d=0}^{D-1}  \Delta I(t_{d}))^{2}}{\sum_{d=0}^{D-1}  I(t_{d})} + \frac{\sum_{d=0}^{D-1}  \Delta I(t_{d})}{\sum_{d=0}^{D-1}  I(t_{d})}    \}.
\label{alphaestimator}
\end{eqnarray}
\begin{eqnarray}
\hat{\beta} = \frac{1}{2 \Delta t} \{ \frac{1}{D} \sum_{d=0}^{D-1}  \frac{\Delta I(t_{d})}{I(t_{d})} - \frac{1}{D} \frac{(\sum_{d=0}^{D-1}  \Delta I(t_{d}))^{2}}{\sum_{d=0}^{D-1}  I(t_{d})} - \frac{\sum_{d=0}^{D-1}  \Delta I(t_{d})}{\sum_{d=0}^{D-1} I(t_{d})}    \}.
\label{betaestimator}
\end{eqnarray}

The second sub-problem is to obtain the maximal likelihood estimator $\hat{\mbox{\boldmath{$l$}}}$ using the obtained values of $\hat{\alpha}$ and $\hat{\beta}$. They are obtained by solving eq.(\ref{gammaestimator}).
\begin{eqnarray}
\hat{\mbox{\boldmath{$l$}}}  = \arg \underset{{\scriptsize \mbox{\boldmath{$l$}}}}{\max} L^{[{\rm I1}]}(\hat{\alpha},\hat{\beta},\Gamma_{ij}(\mbox{\boldmath{$l$}})).
\label{gammaestimator}
\end{eqnarray}

Eq.(\ref{gammaestimator}) can not be solved analytically. There are $10^{14}$ possible topologies for $N=10$, and $10^{57}$ for $N=20$. Simulated annealing \cite{Pre07} is a powerful meta-heuristic algorithm to solve such a combinatorial global optimization problem. A candidate of parameters $\mbox{\boldmath{$l$}}'$ is generated randomly near the present value of $\mbox{\boldmath{$l$}}$. The parameters are updated from $\gamma_{ij}=\Gamma_{ij}(\mbox{\boldmath{$l$}})$ to $\Gamma_{ij}(\mbox{\boldmath{$l$}}')$ according to the probability $p(s)$ in eq.(\ref{SAprob}) in the $s$-th step ($s=0,1,\cdots$) of iterations. 
\begin{eqnarray}
p(s) = \min (\exp(\frac{L^{[{\rm I1}]}(\hat{\alpha},\hat{\beta},\Gamma_{ij}(\mbox{\boldmath{$l$}}))-L^{[{\rm I1}]}(\hat{\alpha},\hat{\beta},\Gamma_{ij}(\mbox{\boldmath{$l$}}')}{k T(s)}),1).
\label{SAprob}
\end{eqnarray}

$T(s)$ is the annealing temperature in the $s$-th step. Typical cooling schedule is $T(s) = 1/\log(s+1)$. Since $O(T)=1$, the scaling constant $k$ is selected as an appropriate value whose order is the same as that of $L^{[{\rm I1}]}$.

\subsubsection{Case 2: for given $\Delta J_{i}(t_{d})$}
\label{estimationfromJ}

The procedure for the estimation from $\Delta J_{i}(t_{d})$ is presented. Again, the problem is divided to two sub-problems. The first sub-problem is to solve eq.(\ref{esalbeJ}). The quantity $I(0)$ is the initial value of the number of infectious persons, which appears in the formula for the mean and variance of $J$ in Appendix \ref{moIJ}. It is not the same as the known $J(t_{0})$, but an unknown parameter.
\begin{eqnarray}
\hat{\alpha}, \ \hat{\beta}, \ \hat{I}(0) = \arg \underset{\alpha,\beta,I(0)}{\max} L^{[{\rm J2}]}(\alpha,\beta,I(0)).
\label{esalbeJ}
\end{eqnarray}

Simulated annealing uses the probability $p(s)$ in eq.(\ref{SAprobJ}) for the update of a candidate. An alternative means to solve eq.(\ref{esalbeJ}) is such a function maximization algorithm as a BFGS quasi-Newton method \cite{Pre07}.
\begin{eqnarray}
p(s) = \min (\exp(\frac{L^{[{\rm J2}]}(\alpha,\beta,I(0))-L^{[{\rm J2}]}(\alpha',\beta',I(0)')}{k T(s)}),1).
\label{SAprobJ}
\end{eqnarray}

A great difficulty in maximizing $L^{[{\rm J1}]}(\mbox{\boldmath{$\theta$}})$ is encountered in solving the second sub-problem. The very complex formula for $\mbox{\boldmath{$v$}}^{{\rm [JJ]}}(t|\mbox{\boldmath{$\theta$}})$ to obtain the value of $L^{[{\rm J1}]}(\mbox{\boldmath{$\theta$}})$ is not tractable even numerically unless $N$ is very small. An approximation is introduced to convert this problem to the computationally tractable second sub-problem in \ref{estimationfromI}. The valued of $I_{i}(t_{d})$ is approximately obtained from the value of $\Delta J_{i}(t_{d})$ by eq.(\ref{JtoI}), which use the already obtained value of $\hat{\alpha}$. Eq.(\ref{gammaestimator}) is solved with the converted values of $I_{i}(t_{d})$ instead of maximizing $L^{[{\rm J1}]}(\mbox{\boldmath{$\theta$}})$ directly. 
\begin{eqnarray}
I_{i}(t_{d}) \approx \frac{\Delta J_{i}(t_{d})}{\hat{\alpha} \Delta t}.
\label{JtoI}
\end{eqnarray}

Eq.(\ref{JtoI}) is a discrete time approximation of eq.(\ref{dJ/dt}) for small $\Delta t$. This relationship holds true for the mean values of $I_{i}(t_{d})$ and $\Delta J_{i}(t_{d})$. But the variance of $I_{i}(t_{d})$ is overestimated by neglecting the stochastic term $\sqrt{\alpha I_{i}} \xi_{i}^{{\rm [\alpha]}}$. Because of the approximation, the estimation from $\Delta J_{i}(t_{d})$ would be more erroneous than that from $I_{i}(t_{d})$. The estimation errors are demonstrated in section \ref{experiment}.

\section{Experiment}
\label{experiment}

\subsection{Computationally synthesized dataset}

A number of test datasets are synthesized by numerical integration \cite{Klo92} of a Langevin equation (\ref{dI/dt}) for random network topologies and transmission parameters. The network is a Erd\"os-R\'enyi model in a combination of $N$ and the average nodal degree $\langle k_{i} \rangle$. The nodal degree of a node $n_{i}$ is given by $k_{i} = \sum_{j=0}^{N-1} l_{ij}$. The probability at which $l_{ij}=1$ is $\langle k_{i} \rangle/(N-1)$.

It is postulated that the total number of persons who moves from $n_{i}$ to $n_{j}$ per a unit time is proportional to $\sqrt{k_{i} k_{j}}$ if a link is present. This law is known valid generally for the world-wide airline transportation network \cite{Bar04}. It is also postulated that the initial population $P_{i}(0)=S_{i}(0)+I_{i}(0)+R_{i}(0)$ of a node $n_{i}$ is proportional to the total number of persons who outgoes from the node per a unit time. Consequently, $\gamma_{ij}$ is determined as a function of $\mbox{\boldmath{$l$}}$ by eq.(\ref{gammacalc}). The fraction of persons who outgoes per a unit time is a constant $\gamma$ over the network. This is an additional unknown parameter in solving eq.(\ref{gammaestimator}). The law in eq.(\ref{gammacalc}) is used in discovering the network topology by eq.(\ref{gammaestimator}) as well as synthesizing the datasets computationally.
\begin{eqnarray}
\gamma_{ij}= \Gamma_{ij}(\mbox{\boldmath{$l$}}) = \frac{l_{ij} \sqrt{k_{i} k_{j}}}{\sum_{j=0}^{N-1} l_{ij} \sqrt{k_{i} k_{j}}} \gamma.
\label{gammacalc}
\end{eqnarray}

$P_{i}(0)$ is given by eq.(\ref{nodepop}). The total population is set to $P=10^{6} N$ in the experiment.
\begin{eqnarray}
P_{i}(0) = \frac{\sum_{j=0}^{N-1} l_{ij} \sqrt{k_{i} k_{j}}}{\sum_{i=0}^{N-1} \sum_{j=0}^{N-1} l_{ij} \sqrt{k_{i} k_{j}}} P.
\label{nodepop}
\end{eqnarray}

The estimation error of the basic reproductive ratio is defined by eq.(\ref{errr}). It is a relative absolute deviation from the true value.
\begin{eqnarray}
E_{r} = \frac{|\hat{r}-r|}{r} = \frac{|\hat{\alpha}/\hat{\beta}-\alpha/\beta|}{\alpha/\beta}.
\label{errr}
\end{eqnarray}

The estimation error of the topology is defined by eq.(\ref{errl}). It is the fraction of links whose presence or absence is estimated wrongly.
\begin{eqnarray}
E_{l} = \frac{\sum_{i < j} |\hat{l}_{ij}-l_{ij}| }{N(N-1)/2}.
\label{errl}
\end{eqnarray}

\begin{figure}
\begin{center}
\includegraphics[scale=0.48,angle=-90]{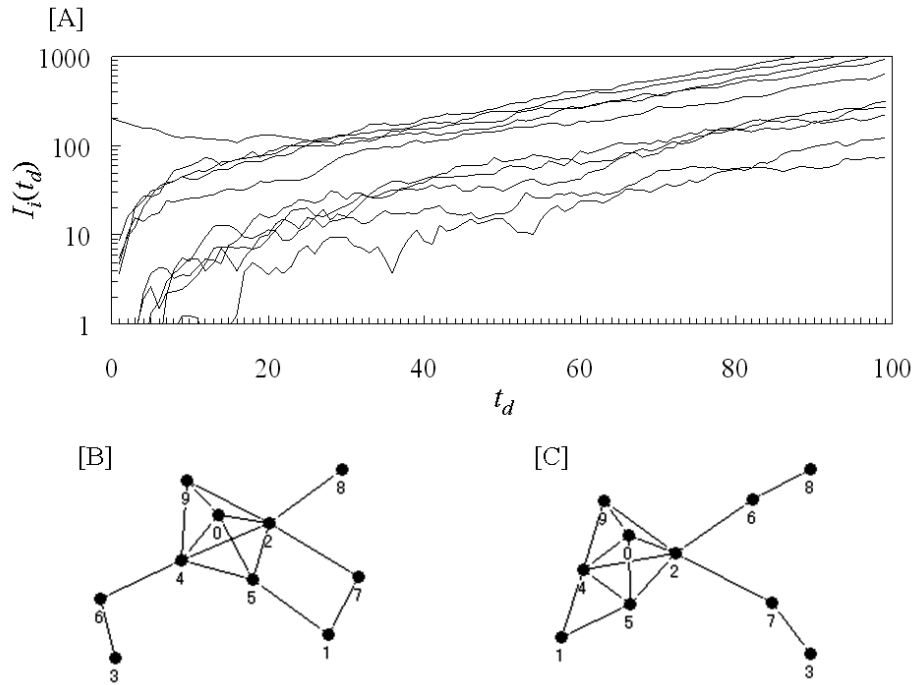}
\end{center}
\caption{Example of the network topology estimated from a computationally synthesized dataset. [A]: dataset $I_{i}(t_{d})$ with $\Delta t=1$ and $D=100$ when the basic reproductive ratio is $r=2 \ (\alpha=0.067, \beta=0.033, \gamma=0.1)$. Individual curves represent the nodes. [B]: random network topology with $N=10$ and $\langle k_{i} \rangle=3$ to synthesize the dataset in [A]. The index cases appear at $n_{0}$. At $t=t_{99}$, $I_{2} > I_{4} > I_{0} > I_{5} > I_{9} > I_{7} > I_{1} > I_{6} > I_{8} > I_{3}$. [C]: network topology estimated from the dataset in [A].}
\label{05313s}
\end{figure}

{\em Figure \ref{05313s} illustrates an example of the network topology estimated from a computationally synthesized dataset. The graph [A] shows the dataset $I_{i}(t_{d})$ with $\Delta t=1$ and $D=100$ when $r=2$. The drawing [B] shows the topology with $N=10$ and $\langle k_{i} \rangle=3$ to synthesize the dataset. The index cases appear at $n_{0}$. The network includes a core sub-structure consisting of $n_{0}$, $n_{2}$, $n_{4}$, $n_{5}$, and $n_{9}$. It is nearly a clique where every node is connected to every other node. Links are present except for the one between $n_{5}$ and $n_{9}$. The drawing [C] shows the topology estimated from the dataset. The error is $E_{l}=0.18$. The core is discovered correctly. The ability of the method is surprising in distinguishing the only pair of nodes where the link is absent. The links from the core to $n_{1}$ and $n_{7}$ are discovered. Although the method identifies that $n_{3}$, $n_{6}$, and $n_{8}$ do not belong to the core, but form the stubs (dead ends) from the core, it fails to estimate how they are connected to each other and the core. The number of cases is the smallest at these nodes. The movements of infectious persons to and from them are so infrequent that the analysis on them is not so reliable.}

\begin{figure}
\begin{center}
\includegraphics[scale=0.34,angle=-90]{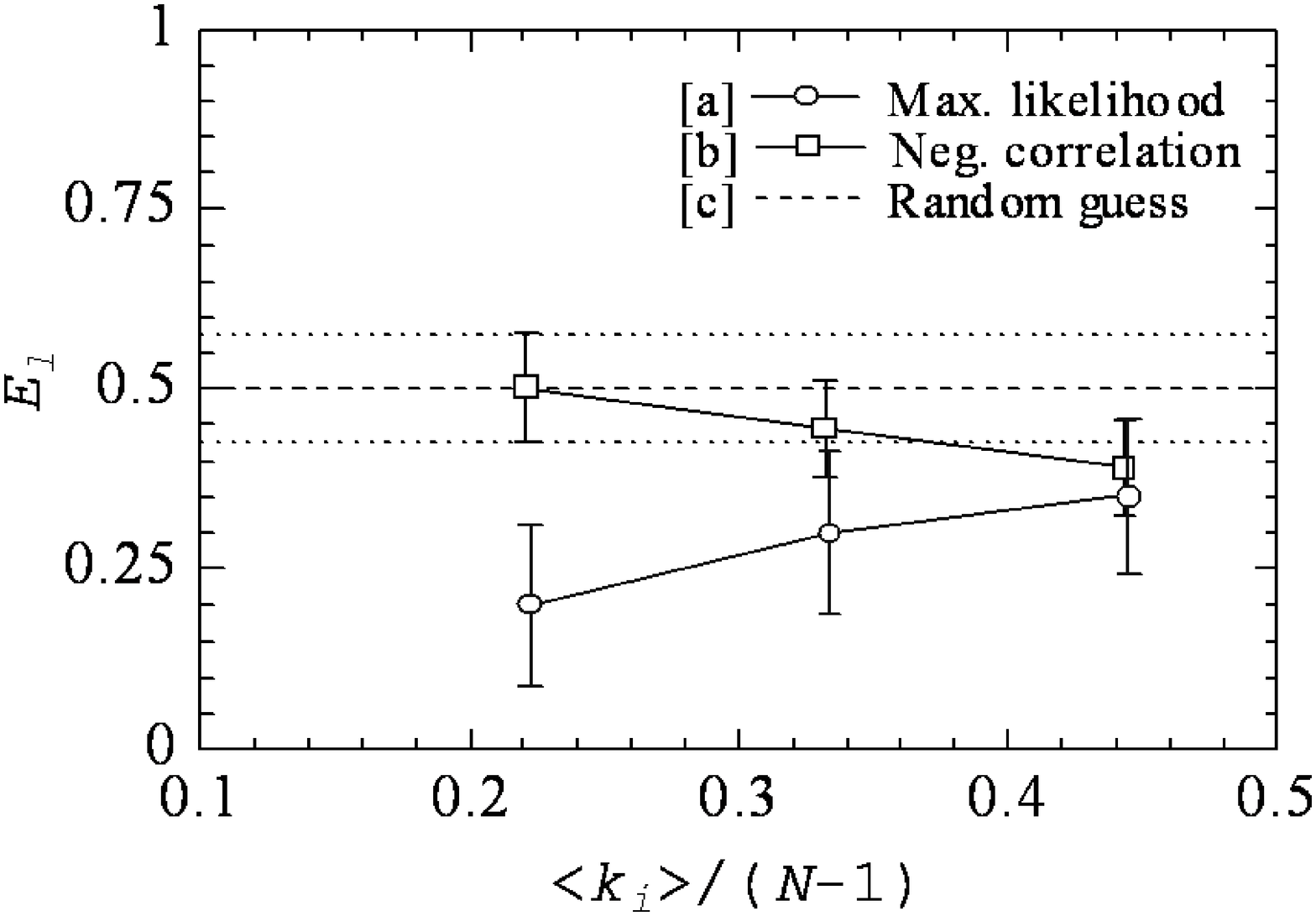}
\end{center}
\caption{Estimation error $E_{l}$ for various values of the normalized average degree $\langle k_{i} \rangle/(N-1)$ ($\langle k_{i} \rangle=2,3,4$ for the number of nodes $N=10$) when $r=2 \ (\alpha=0.067, \beta=0.033, \gamma=0.1)$, and $I_{i}(t_{d})$ with $\Delta t=1$ and $D=100$ is given as a dataset. The initial condition is $I_{0}(0)=200$ and $I_{i}(0)=0$ for all $i \neq 0$. The individual plots show the mean and standard deviation over trials for 100 different random networks. [a]: maximal likelihood estimation (the method presented in this study). [b]: naive negative correlation estimation. [c]: mere random guess (theoretically $0.5 \pm 0.0745$).}
\label{05311s}
\end{figure}

{\em The estimation error $E_{l}$ of the method in this study is compared with those of a naive estimation and a mere random guess. The naive estimation relies on the correlation between nodes. When an infectious person moves from $n_{i}$ to $n_{j}$, $I_{i}$ decreases and $I_{j}$ increases by one simultaneously. Thus, intuitively, the negative correlation of the change in $I_{i}$ and $I_{j}$ is the signal of the presence of a link. The correlation $\rho_{ij}$ between $n_{i}$ and $n_{j}$ is calculated by eq.(\ref{correl}) where $\Delta I_{i}(t_{d})=I_{i}(t_{d+1}) - I_{i}(t_{d})$.
\begin{eqnarray}
\rho_{ij} = \sum_{d=0}^{D-2} (\Delta I_{i}(t_{d}) - \frac{1}{N} \sum_{k=0}^{N-1} \Delta I_{k}(t_{d})) (\Delta I_{j}(t_{d}) - \frac{1}{N} \sum_{k=0}^{N-1} \Delta I_{k}(t_{d})).
\label{correl}
\end{eqnarray}

Note that $\sum_{k} \Delta I_{k=0}^{N-1}(t_{d})/N$ is not the average over the time sequence for $n_{k}$, but the average over the nodes at $t=t_{d}$. This formula is supposed to exclude the positive correlation because of the common growing trends in $I_{i}$ (dependent on $(\alpha-\beta)I_{i} \Delta t$ in eq.(\ref{apmIisol})). The naive estimation predicts $l_{ij} =1$ if $\rho_{ij}<0$. The random guess is the worst bound of estimation. The number of links whose presence or absence is predicted wrongly obeys a binomial distribution. The mean and standard deviation of $E_{l}$ are 0.5 and 0.0745 for $N=10$, and 0.5 and 0.0256 for $N=20$ theoretically. 

Figure \ref{05311s} shows $E_{l}$ for various values of the normalized average degree $\langle k_{i} \rangle/(N-1)$ ($\langle k_{i} \rangle=2,3,4$ for the number of nodes $N=10$) when $r=2$, and $I_{i}(t_{d})$ with $\Delta t=1$ and $D=100$ is given as a dataset. For small $\langle k_{i} \rangle $, the naive estimation does not work at all. As $\langle k_{i} \rangle $ increases, $E_{l}$ of the method increases and that of the naive estimation decreases. For large $\langle k_{i} \rangle$, the naive estimation becomes less erroneous than the random guess. But, it never surpasses the method in this study. The initially heterogeneous node-to-node distribution of infectious persons relaxes more quickly in the networks having more links. For example, the standard deviation of $I_{i}(t_{99})$ is about 400 for $\langle k_{i} \rangle=2$, and about 300 for $\langle k_{i} \rangle=4$, while the mean for both $\langle k_{i} \rangle$ is about 550 ($\approx I_{0}(0) \exp ((\alpha-\beta)D \Delta t)/N$). The growing trends also become more homogeneous. Under such homogeneity, the negative correlation implies the movements between nodes directly. The naive estimation may be a substitute if assuming homogeneous distribution during observation are well-grounded.}

\begin{figure}
\begin{center}
\includegraphics[scale=0.62,angle=0]{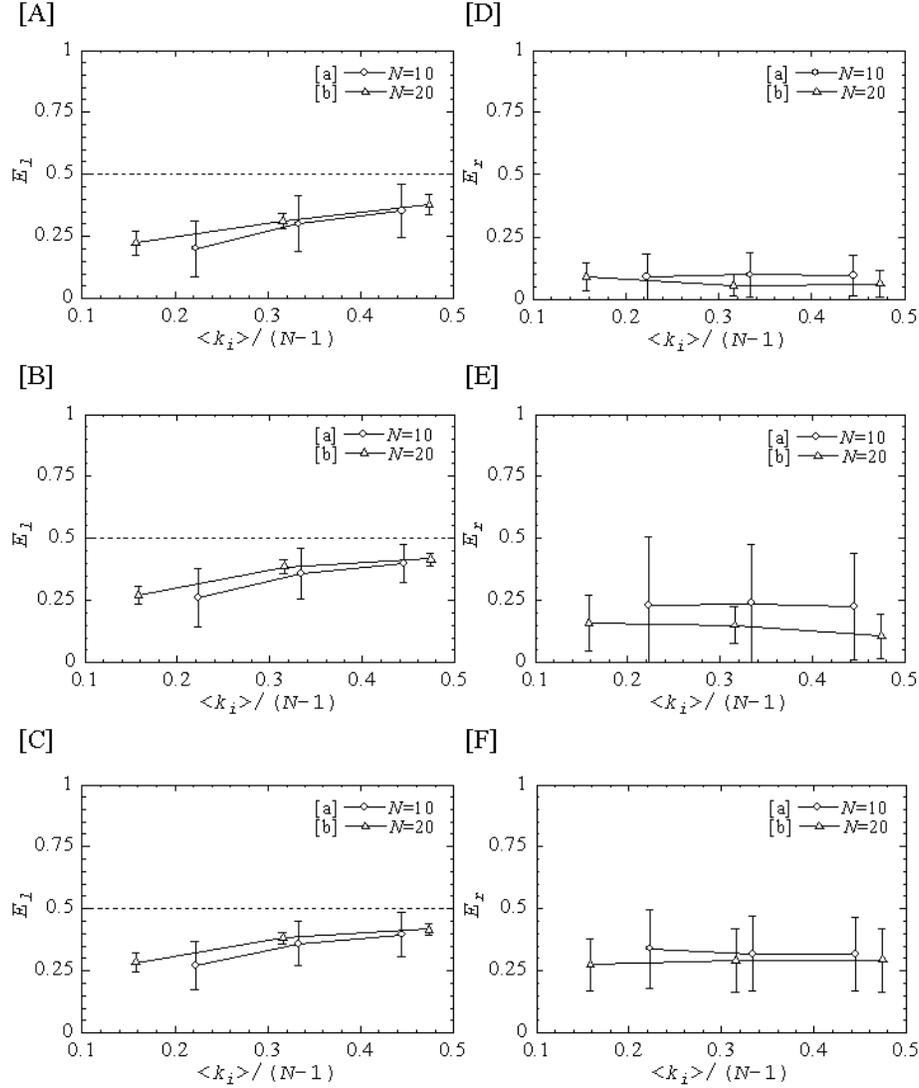}
\end{center}
\caption{Estimation errors $E_{l}$ and $E_{r}$ for various values of the normalized average degree $\langle k_{i} \rangle/(N-1)$ ($\langle k_{i} \rangle=2,3,4$ for $N=10$ and $\langle k_{i} \rangle=3,6,9$ for $N=20$), the number of nodes $N$, and the basic reproductive ratio $r$ when $I_{i}(t_{d})$ with $\Delta t=1$ and $D=100$ is given as a dataset. The initial condition is $I_{0}(0)=200$ and $I_{i}(0)=0$ for all $i \neq 0$. The individual plots show the mean and standard deviation over trials for 100 different random networks. [A]: $E_{l}$ for $r=2$ ($\alpha=0.067$, $\beta=0.033$, $\gamma=0.1$). [B]: $E_{l}$ for $r=4$ ($\alpha=0.08$, $\beta=0.02$, $\gamma=0.1$). [C]: $E_{l}$ for $r=6$ ($\alpha=0.086$, $\beta=0.014$, $\gamma=0.1$). [D]: $E_{r}$ for $r=2$ (the same as [A]). [E]: $E_{r}$ for $r=4$ (the same as [B]). [F]: $E_{r}$ for $r=6$ (the same as [C]).}
\label{0303a1s}
\end{figure}

Figure \ref{0303a1s} shows the estimation errors $E_{l}$ and $E_{r}$ for various values of the normalized average degree $\langle k_{i} \rangle/(N-1)$ ($\langle k_{i} \rangle=2,3,4$ for $N=10$ and $\langle k_{i} \rangle=3,6,9$ for $N=20$), $N$, and $r$ when $I_{i}(t_{d})$ with $\Delta t=1$ and $D=100$ is given as a dataset. The findings are as follows.

\begin{itemize}
\item As $\langle k_{i} \rangle$ increases, $E_{l}$ increases from around 0.2 to 0.4. Although $E_{l}$ (the average $\pm$ standard deviation) remains less than $0.5$ within the range of the experimental conditions here, the estimation comes close to a mere random guess ($E_{l} \rightarrow 0.5$) for the dense network limit ($\rightarrow$ a complete graph). Discerning the presence or absence of links becomes more difficult as the spread goes on over more links in parallel and reaches more nodes along more possible routes. On the other hand, $E_{r}$ does not change largely as $\langle k_{i} \rangle$ changes. 
\item As the network becomes larger, $E_{l}$ increases and $E_{r}$ decreases. As the model becomes more complex (the number of links $O(N^{2})$ becomes larger compared to the amount of data $O(ND)$), there may appear more optimal or sub-optimal topologies. Choosing a unique right answer becomes more difficult from such similar candidates. On the other hand, the central limit theorem guarantees that the fluctuation decreases as the network becomes larger because $\alpha$ and $\beta$ are estimated from the sum of $N$ probability variables ($I(t_{d})=\sum_{i=0}^{N-1} I_{i}(t_{d})$).
\item As $r$ increases, $E_{l}$ increases (but the difference between $r=4$ and $r=6$ is very small) and $E_{r}$ increases from around 0.1 to 0.35. The observations can not track down the rapid reproduction of patients when $r \Delta t$ is large. 
\end{itemize}

{\em The dependence of the errors on $\gamma$ in eq.(\ref{gammacalc}) is investigated. The errors increase from $E_{l}=0.2$ and $E_{r}=0.092$ for $\gamma=0.1$ in [a] of figure \ref{0303a1s} [A] to $E_{l}=0.26$ and $E_{r}=0.10$ for $\gamma=0.2$, and $E_{l}=0.32$ and $E_{r}=0.099$ for $\gamma=0.4$. The accuracy of estimation is limited when many persons move between nodes in both directions because of large $\gamma$. The dependence of the errors on $\Delta t$ is investigated. The errors increase from $E_{l}=0.2$ and $E_{r}=0.092$ in [a] of figure \ref{0303a1s} [A] to $E_{l}=0.31$ and $E_{r}=0.18$ if the observations are made four times less frequently ($\Delta t=4$, $D=25$). But they are improved only slightly to $E_{l}=0.18$ and $E_{r}=0.074$ if the observations are made 4 times more frequently ($\Delta t=0.25$, $D=400$). Small time interval between observations is relevant to accurate estimation. The errors are investigated for various initial population distributions $P_{i}(0)$. If the population is a thousandth ($P=10^{3}N$), $E_{l}=0.23$ and $E_{r}=0.091$ for $N=10$, $\langle k_{i} \rangle=2$, and $r=2$ when $I_{0}(0)=20$. If $P_{i}(0) \propto \sum_{j=0}^{N-1} l_{ij} (k_{i}k_{j})^{4}$ rather than $\sum_{j=0}^{N-1} l_{ij} \sqrt{k_{i}k_{j}}$ in eq.(\ref{nodepop}), $E_{l}=0.25$ and $E_{r}=0.09$ when the population is $P=10^{6}N$. In this case, the population ranges in vastly diverse scales. The ratio of the population of the most populated node to that of the least populated node is $P_{{\rm max}}(0)/P_{{\rm min}}(0) \approx 2000$ while $P_{{\rm max}}(0)/P_{{\rm min}}(0) \approx 7$ in case of eq.(\ref{nodepop}). $E_{r}$ is not affected by the distribution. $E_{l}$ increases when much less populated nodes are present. But $E_{l}$ still remains small.}

\begin{figure}
\begin{center}
\includegraphics[scale=0.62,angle=0]{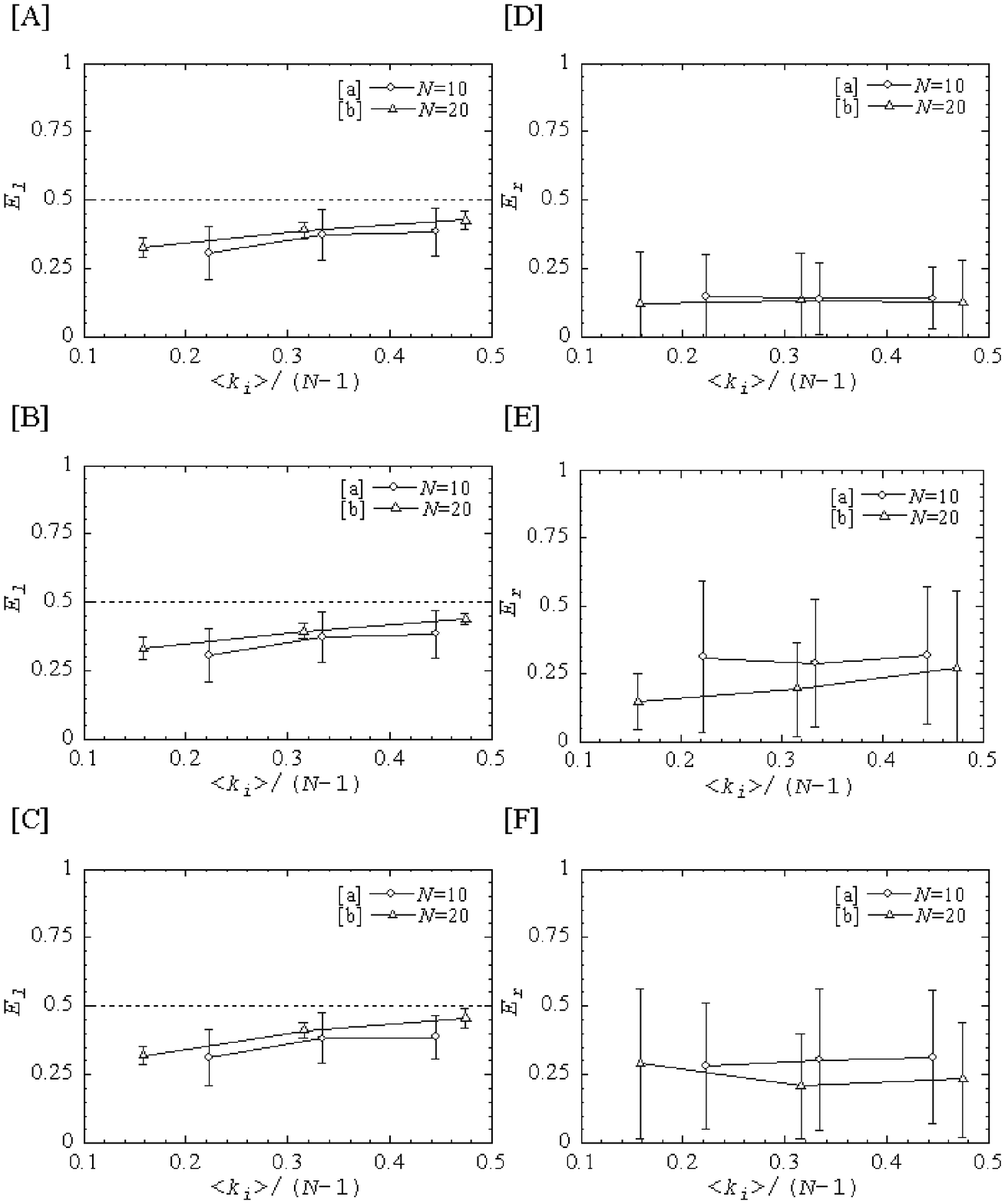}
\end{center}
\caption{Estimation errors $E_{l}$ and $E_{r}$ for various values of $\langle k_{i} \rangle/(N-1)$, $N$, and $r$ when $\Delta J_{i}(t_{d})$ with $\Delta t=1$ and $D=100$ is given as a dataset. The experimental conditions are the same as those for Figure \ref{0303a1s}.}
\label{0303a2s}
\end{figure}

Figure \ref{0303a2s} shows the estimation errors $E_{l}$ and $E_{r}$ for various values of $\langle k_{i} \rangle/(N-1)$, $N$, and $r$ when $\Delta J_{i}(t_{d})$ with $\Delta t=1$ and $D=100$ is given as a dataset. The experimental conditions are the same as those for Figure \ref{0303a1s}. The findings are as follows.

\begin{itemize}
\item The dependency of $E_{l}$ and $E_{r}$ on $\langle k_{i} \rangle$, $N$, and $r$ in figure \ref{0303a2s} is similar to those in figure \ref{0303a1s}.
\item The absolute value of errors tends to increase. For example, $E_{l}=0.31$ in figure \ref{0303a2s} is much larger than $E_{l}=0.2$ in figure \ref{0303a1s} under the same experimental conditions $\langle k_{i} \rangle=2$, $N=10$, and $r=2$. In contrast, the increase in $E_{r}$ is relatively small. $E_{r}$ is nearly the same when $r=6$. The deterioration in estimating the topology, therefore, seem to result from the influence of the approximation in eq.(\ref{JtoI}). 
\end{itemize}

As a summary of the experiments, the estimation is particularly reliable ($E_{l} \sim 0.2$ and $E_{r} \sim 0.1$) when $I_{i}(t_{d})$ for a slow reproduction over a sparse network is used as an input. Such performance can not be achieved by the naive estimation. The estimated topology from $\Delta J_{i}(t_{d})$ is more erroneous than that from $I_{i}(t_{d})$ by as much as 50\%.

\subsection{SARS dataset}

SARS is a respiratory disease in humans caused by the SARS corona-virus. The epidemic of SARS appears to have started in Guangdong Province of south China in November 2002. SARS spread from the Guangdong Province to Hong Kong in early 2003, and eventually nearly 40 countries around the world by July. WHO archives the cumulative number of reported probable cases of SARS\footnote{World Health Organization, Cumulative number of reported probable cases of SARS, http://www.who.int/csr/sars/country/en/index.html (2003).}. The dataset in the archive had been updated nearly every day since March 17. It is a time sequence dataset $J_{i}(t_{d})$ with $\Delta t=1$ day. In this study, the target geographical regions are those where five or more cases had been reported in a month since March 17. They include Canada (CAN), France (FRA), United Kingdom (GBR), Germany (GER), Hong Kong (HKG), Malaysia (MAS), Taiwan (ROC), Singapore (SIN), Thailand (THA), United States (USA), and Vietnam (VIE). Mainland China is not included because no data is available in some periods and no data outside of Guangdong Province is reported in other periods.

\begin{figure}
\begin{center}
\includegraphics[scale=0.38,angle=-90]{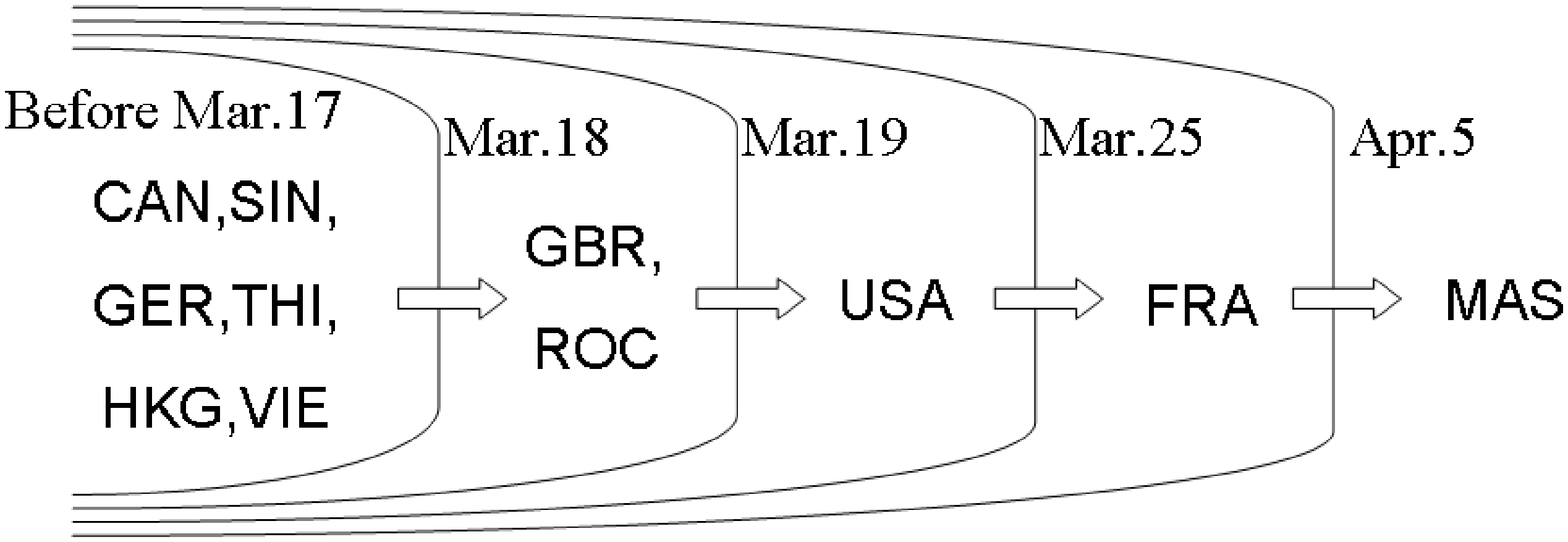}
\end{center}
\caption{Date when the first cases appeared and the propagating wavefront of the spread in the WHO dataset on the cases of SARS.}
\label{05312s}
\end{figure}

{\em Figure \ref{05312s} shows the date when the first patient appeared and the propagating wavefront of the spread. It is almost certain that neither FRA nor MAS are the origin of the outbreak. But, nobody can tell the chain of transmission among CAN, HKG, SIN, GBR, ROC, and USA in just two days from March 17 to 19 reliably. The wavefront is not as informative as anticipated. Such a naive gleaning verifies the obvious series of events at best.}

The estimated transmission parameters are $\hat{\alpha}=0.18$ and $\hat{\beta}=0.13$. The basic reproductive ratio is $\hat{r}=1.4$. According to the field-based medical case studies, the basic reproductive ratio (except for super-spreading events \cite{Fuj07}) was $r=2.7$ in February and went down to $r=1$ in late March in Hong Kong \cite{Ril03}, and $r=7$ in February and $r=1$ in early March in Singapore \cite{Lip03}. The decrease of $r$ is due to the quarantine, hospitalization and public awareness starting to take effects after WHO issued a world-wide alert on March 12. But the spread of SARS was still going on world-wide. The value slightly greater than 1 seems reasonable at that stage.

\begin{figure}
\begin{center}
\includegraphics[scale=0.4,angle=-90]{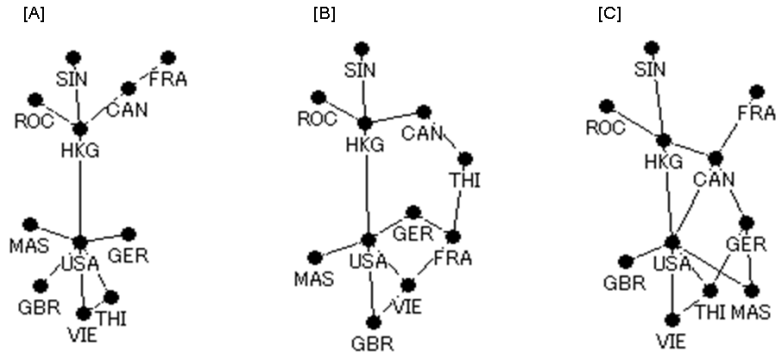}
\end{center}
\caption{Estimated topologies from the WHO dataset on the cases of SARS in Canada (CAN), France (FRA), United Kingdom (GBR), Germany (GER), Hong Kong (HKG), Malaysia (MAS), Taiwan (ROC), Singapore (SIN), Thailand (THI), United States (USA), and Vietnam (VIE) from March 17 through April 17. [A]: the most likely topology. [B]: the second most likely topology. [C]: the third most likely topology.}
\label{0303a3s}
\end{figure}

It is postulated that the law in eq.(\ref{gammacalc}) holds true in analyzing the SARS dataset. The topology which achieves the largest value of the logarithmic likehood function among many trials is chosen. This is efficient in rejecting the local maximum to which simulated annealing may converge. The trials use different random number sequences to generate nearby candidates $\mbox{\boldmath{$l$}}'$ in eq.(\ref{SAprob}). Figure \ref{0303a3s} shows the estimated topologies $\hat{\mbox{\boldmath{$l$}}}$. The topology [A] is the most likely (the largest value of the likelihood $L=-9985$). The best 30 trials out of 300 trials converge to [A]. It includes 11 links ($\langle k_{i} \rangle/(N-1) =0.2$). The topology [B] is the second most likely ($L=-9998$). The next 5 trials converge to [B]. It include 13 links ($\langle k_{i} \rangle/(N-1) =0.24$). The topology [C] is the third most likely ($L=-10012$). The next 29 trials converge to [C]. It include 14 links ($\langle k_{i} \rangle/(N-1) =0.25$). About 20\% of the trials converge to either [A] or [C]. 

The sub-structures common in all of [A], [B], and [C] are a star from HKG to CAN, ROC, and SIN, another star from USA to GBR, MAS, and VIE, and a link between the centers of these stars (HKG and USA). A triangle between USA, VIE, and THI appears in [A] and [C]. {\em The likelihood seems sensitive to the topological whereabouts of GER and FRA given these common sub-structures as a core of the network. This may happen to make [B] a tall but narrow peak in the landscape of the likelihood, which simulated annealing sometimes fail to discover.} In addition to these sub-structures, a few remarkable points are seen in these topologies. The nodes SIN and ROC are stubs where $k_{i}=1$. The role of SIN is not so relevant in spreading the disease despite the fact that the number of cases there was more than 100 in the middle of April. The nodes CAN and USA have links to distant geographical regions, and USA is a hub ($k_{i}$ is the largest). They are relevant intermediate spreader nodes. The links around GER are not stable among the three topologies. The number of cases in some European countries is too small to draw a reliable conclusion.

The estimated topologies are not meant to reproduce the trajectories of individual patients' movement, but rather demonstrate some demographical interactions within the macroscopic world-wide transportation behind the SARS outbreak. Nevertheless, the sub-structures mentioned above seem to be consistent with the following publicly known series of events on some individual patients' microscopic movements.

\begin{itemize}
\item Two of the index patients in Toronto in Canada, three of the index patients in Singapore, and another three of the index patients in the United States stayed a hotel in Hong Kong where a Chinese nephrologist, who had treated many patients in Guangzhou and become infected, was staying in late February\footnote{SARS Expert Committee (Hong Kong), SARS in Hong Kong: from experience to action, http://www.sars-expertcom.gov.hk/english/reports/reports/reports\_fullrpt.html (2003).}. This event implies the links from HKG to CAN, SIN, and USA form a chain of transmission in the early growth phase of SARS outbreak.
\item A garment manufacturer from the United States became infected during the stay in Hong Kong on the way to Hanoi in Vietnam, showed symptoms there, and was evacuated to a hospital in Hong Kong \cite{Gre06}. An Italian physician, who treated him at a hospital in Hanoi, showed symptoms in Bangkok in Thailand where he would attend a conference in early March. These events imply that the interactions among HKG, USA, VIE, and THI are present potentially, which could result in another chain of transmission.
\end{itemize}

The WHO dataset is not of perfectly reliable quality. Particularly, the data on mainland China is of poor quality, and can not be used in this study. Even the individual number of cases which was reported from the other local governments may not be accurate. The number of cases is highly fluctuating and seems noisy. Data in a city-level resolution, rather than nation-level, would be necessary for accurate estimation when large countries like USA play an important role as a spreader. It is surprising that, in spite of these limitations, the method reproduces some characteristics of the network over which SARS spread from Hong Kong to Southeast Asia and North America.

\section{Conclusion}

The method presented in this study solves an inverse problem to discover the effectively decisive network and reveal the transmission parameters from the observation $I_{i}(t_{d})$ or $\Delta J_{i}(t_{d})$ on the spread of an infectious disease. The findings with test datasets are that the estimation is particularly successful when the topology is sparse and reproduction is slow, and that the estimation from $\Delta J_{i}(t_{d})$ is more erroneous than that from $I_{i}(t_{d})$. The network topology discovered from a seemingly noisy dataset on the SARS outbreak reproduces some characteristic patterns of the spread from Hong Kong to Southeast Asia and North America. So far, a great effort has been made to get a complete picture of how an infectious disease did and will spread from the found pieces of an epidemiological jigsaw puzzle. The method presents new pieces from a viewpoint of macroscopic transportation. These pieces can be put together with the pieces found in the conventional field-based medical case studies on the individual patients' microscopic movements.

The method can be extended to apply to a more practical situation. The experimental condition in this study is an extreme where nothing but $I_{i}(t_{d})$ or $\Delta J_{i}(t_{d})$ is known and no informative prior knowledge is available. If some demographical statistics on the traffic between cities or the findings on the past contacts between individual patients are available, the consequent posterior distribution enables more comprehensive Bayesian inference. Another extension is to employ more complicated but realistic epidemiological compartment models. Latent period (infected but not infectious) and hospitalization are relevant for some diseases. The dependence of recovery on time ($\beta \neq$ constant) is realistic for other diseases. {\em Strictly speaking, the time interval from infection to recovery and from a movement to another obeys appropriate probability density functions.  Analytical treatment of the stochastic process with these effects tends to be considerably difficult. The estimation may count on such a numerical method as a Markov-chain Monte-Carlo sampling.

In addition to the extensions to the method, it is sometimes critical to gather such a larger dataset as a collection of multiple independent time sequences starting from different index cases. Less erroneous estimation may be possible even for the regions whose population is small or where the number of cases is small. Such a dataset is not available for SARS, but possibly for influenza which spreads around the world in seasonal epidemics.} Understanding the landscape of the likelihood functions is essential in identifying the requisites for a dataset, a network topology, and a mathematical model of disease transmission to make the inverse problem well-posed and stabilize the solution. This remains the challenge for the future.

\appendix

\section{Probability density function}
\label{Solving}

A generic form of a Langevin equation for multiple time-dependent variables $x_{i}(t)$ is given by eq.(\ref{NvarSLE}). The fluctuations $\xi_{a}(t)$ are stochastic terms.
\begin{eqnarray}
\frac{{\rm d} x_{i}(t)}{{\rm d}t} = \mu_{i}(x_{0}(t),\cdots,x_{N-1}(t)) + \sum_{a=0}^{M-1} \sigma_{ia}(x_{0}(t),\cdots,x_{N-1}(t)) \xi_{a}(t).
\label{NvarSLE}
\end{eqnarray}

Eq.(\ref{NvarSLE}) can be solved by deriving the probability density function $p(\mbox{\boldmath{$x$}},t)$ for probability variables $\mbox{\boldmath{$x$}}=(x_{0},\cdots,x_{N-1})$ at time $t$. The time evolution of $p(\mbox{\boldmath{$x$}},t)$ is given by the Fokker-Planck equation in eq.(\ref{NvarFPE}).
\begin{eqnarray}
\frac{\partial p(\mbox{\boldmath{$x$}},t)}{\partial t} = -\sum_{i=0}^{N-1} \frac{\partial}{\partial x_{i}} A_{i}(\mbox{\boldmath{$x$}}) p(\mbox{\boldmath{$x$}},t) + \frac{1}{2} \sum_{i,j=0}^{N-1} \frac{\partial^{2}}{\partial x_{i} \partial x_{j}} B_{ij}(\mbox{\boldmath{$x$}}) p(\mbox{\boldmath{$x$}},t).
\label{NvarFPE}
\end{eqnarray}

The coefficiets $A_{i}$ and $B_{ij}$ are given by eq.(\ref{Adef}) and (\ref{Bdef}).
\begin{eqnarray}
A_{i}(\mbox{\boldmath{$x$}}) = \mu_{i}(\mbox{\boldmath{$x$}}).
\label{Adef}
\end{eqnarray}
\begin{eqnarray}
B_{ij}(\mbox{\boldmath{$x$}}) = \sum_{a=0}^{M-1} \sigma_{ia}(\mbox{\boldmath{$x$}}) \sigma_{ja}(\mbox{\boldmath{$x$}}).
\label{Bdef}
\end{eqnarray}

The mean (the first order moment) of $x_{i}$ at $t$ is given by $m_{i}(t) = \langle x_{i} \rangle_{t} = \int x_{i} p(\mbox{\boldmath{$x$}},t) {\rm d}\mbox{\boldmath{$x$}}$. The time evolution of $m_{i}(t)$ is given by eq.(\ref{mevo}). It is derived by multiplying eq.(\ref{NvarFPE}) by $x$ and partial integration under the condition where $p$ and $\partial p/\partial x$ decay more rapidly than $A_{i}$ and $B_{ij}$ near the boundary of the domain of $x$.
\begin{eqnarray}
\frac{{\rm d} m_{i}(t)}{{\rm d}t} = \langle A_{i}(\mbox{\boldmath{$x$}}) \rangle_{t}.
\label{mevo}
\end{eqnarray}

The covariance (the second order moment) between $x_{i}$ and $x_{j}$ at $t$ is given by $v_{ij}(t) = \langle x_{i} x_{j} \rangle_{t}-m_{i}(t) m_{j}(t)$. The time evolution of $v_{ij}(t)$ is given by eq.(\ref{vevo}). Derivation is similar to that for eq.(\ref{mevo}).
\begin{eqnarray}
\frac{{\rm d} v_{ij}(t)}{{\rm d}t} = \langle B_{ij}(\mbox{\boldmath{$x$}}) \rangle_{t} + \langle x_{i} A_{j}(\mbox{\boldmath{$x$}}) \rangle_{t} + \langle A_{i}(\mbox{\boldmath{$x$}}) x_{j} \rangle_{t}.
\label{vevo}
\end{eqnarray}

Higher order moments can be obtained recursively as a solution of the differential equations which include the calculated lower order moments.

\section{Moments of $I_{i}$ and $J_{i}$}
\label{moIiJi}

Eq.(\ref{mIievo}) through (\ref{vJJijevo}) are the differential equations for the time evolution of the first and second order moments of $I_{i}$ and $J_{i}$. The symbols $\mbox{\boldmath{$m$}}^{{\rm [I]}}(t|\mbox{\boldmath{$\theta$}})$, $\mbox{\boldmath{$m$}}^{{\rm [J]}}(t|\mbox{\boldmath{$\theta$}})$ are the row vectors whose $i$-th element is the mean of $I_{i}$, $J_{i}$, and $\mbox{\boldmath{$v$}}^{{\rm [II]}}(t|\mbox{\boldmath{$\theta$}})$, $\mbox{\boldmath{$v$}}^{{\rm [IJ]}}(t|\mbox{\boldmath{$\theta$}})$, $\mbox{\boldmath{$v$}}^{{\rm [JJ]}}(t|\mbox{\boldmath{$\theta$}})$ are the $N \times N$ matrices whose $i$-th row and $j$-th column element is the covariance between $I_{i}$ and $I_{j}$, $I_{i}$ and $J_{j}$, $J_{i}$ and $J_{j}$. The unknown network topology and transmission parameters are represented by a symbol $\mbox{\boldmath{$\theta$}}=\{\mbox{\boldmath{$\gamma$}}, \alpha, \beta \}$.
\begin{eqnarray}
\frac{{\rm d} \mbox{\boldmath{$m$}}^{{\rm [I]}}(t|\mbox{\boldmath{$\theta$}})}{{\rm d}t} = \mbox{\boldmath{$m$}}^{{\rm [I]}}(t|\mbox{\boldmath{$\theta$}}) \mbox{\boldmath{$a$}}^{{\rm T}}.
\label{mIievo}
\end{eqnarray}
\begin{eqnarray}
\frac{{\rm d}\mbox{\boldmath{$m$}}^{{\rm [J]}}(t|\mbox{\boldmath{$\theta$}})}{{\rm d}t} = \alpha \mbox{\boldmath{$m$}}^{{\rm [I]}}(t|\mbox{\boldmath{$\theta$}}).
\label{mJievo}
\end{eqnarray}
\begin{eqnarray}
\frac{{\rm d} \mbox{\boldmath{$v$}}^{{\rm [II]}}(t|\mbox{\boldmath{$\theta$}})}{{\rm d} t} = \mbox{\boldmath{$a$}} \mbox{\boldmath{$v$}}^{{\rm [II]}}(t|\mbox{\boldmath{$\theta$}}) + \mbox{\boldmath{$v$}}^{{\rm [II]}}(t|\mbox{\boldmath{$\theta$}}) \mbox{\boldmath{$a$}}^{{\rm T}} + \langle \mbox{\boldmath{$B$}} \rangle_{t}.
\label{vIIijevo}
\end{eqnarray}
\begin{eqnarray}
\frac{{\rm d} \mbox{\boldmath{$v$}}^{{\rm [IJ]}}(t|\mbox{\boldmath{$\theta$}})}{{\rm d} t} = \mbox{\boldmath{$a$}} \mbox{\boldmath{$v$}}^{{\rm [IJ]}}(t|\mbox{\boldmath{$\theta$}}) + \alpha (\mbox{\boldmath{$v$}}^{{\rm [II]}}(t|\mbox{\boldmath{$\theta$}}) + \mbox{\boldmath{$c$}}(t)).
\label{vIJijevo}
\end{eqnarray}
\begin{eqnarray}
\frac{{\rm d} \mbox{\boldmath{$v$}}^{{\rm [JJ]}}(t|\mbox{\boldmath{$\theta$}})}{{\rm d} t} = \alpha (\mbox{\boldmath{$v$}}^{{\rm [IJ]}}(t|\mbox{\boldmath{$\theta$}}) + \mbox{\boldmath{$v$}}^{{\rm [IJ]}}(t|\mbox{\boldmath{$\theta$}})^{{\rm T}} + \mbox{\boldmath{$c$}}(t)).
\label{vJJijevo}
\end{eqnarray}

Definitions of the $N \times N$ matrices $\mbox{\boldmath{$a$}}$, $\mbox{\boldmath{$B$}}$, and $\mbox{\boldmath{$c$}}$ which appear in eq.(\ref{mIievo}) through (\ref{vJJijevo}) are given by eq.(\ref{SIRaij}) through (\ref{SIRcij}).
\begin{eqnarray}
a_{ij} = (\alpha-\beta-\sum_{k=0}^{n-1} \gamma_{ik})\delta_{ij} + \gamma_{ji}.
\label{SIRaij}
\end{eqnarray}
\begin{eqnarray}
B_{ij} = \{(\alpha+\beta+\sum_{k=0}^{N-1} \gamma_{ik}) I_{i}+\sum_{k=0}^{N-1} \gamma_{ki} I_{k} \} \delta_{ij} - \gamma_{ij} I_{i} - \gamma_{ji} I_{j}.
\label{SIRBij}
\end{eqnarray}
\begin{eqnarray}
c_{ij}(t) = \delta_{ij} m^{{\rm [I]}}_{i}(t|\mbox{\boldmath{$\theta$}}).
\label{SIRcij}
\end{eqnarray}

Eq.(\ref{mIisol}) through (\ref{vJJijsol}) are the solutions. $\mbox{\boldmath{$E$}}$ is a unit matrix.
\begin{eqnarray}
\mbox{\boldmath{$m$}}^{{\rm [I]}}(t|\mbox{\boldmath{$\theta$}}) = \mbox{\boldmath{$I$}}(0) \exp (\mbox{\boldmath{$a$}}^{{\rm T}} t). 
\label{mIisol}
\end{eqnarray}
\begin{eqnarray}
\mbox{\boldmath{$m$}}^{{\rm [J]}}(t|\mbox{\boldmath{$\theta$}}) = \mbox{\boldmath{$I$}}(0) \{ \alpha (\mbox{\boldmath{$a$}}^{{\rm T}})^{-1} \exp (\mbox{\boldmath{$a$}}^{{\rm T}} t) - \alpha (\mbox{\boldmath{$a$}}^{{\rm T}})^{-1} +\mbox{\boldmath{$E$}} \}.
\label{mJisol}
\end{eqnarray}
\begin{eqnarray}
\mbox{\boldmath{$v$}}^{{\rm [II]}}(t|\mbox{\boldmath{$\theta$}}) = \int_{0}^{t} \exp (\mbox{\boldmath{$a$}} (t-t')) \ \langle \mbox{\boldmath{$B$}} \rangle_{t'} \ \exp (\mbox{\boldmath{$a$}}^{{\rm T}} (t-t')) \ {\rm d} t'.
\label{vIIijsol}
\end{eqnarray}
\begin{eqnarray}
\mbox{\boldmath{$v$}}^{{\rm [IJ]}}(t|\mbox{\boldmath{$\theta$}}) = \int_{0}^{t} \alpha \exp (\mbox{\boldmath{$a$}} (t-t')) \ (\mbox{\boldmath{$v$}}^{{\rm [II]}}(t'|\mbox{\boldmath{$\theta$}}) + \mbox{\boldmath{$c$}}(t')) \ {\rm d} t'.
\label{vIJijsol}
\end{eqnarray}
\begin{eqnarray}
\mbox{\boldmath{$v$}}^{{\rm [JJ]}}(t|\mbox{\boldmath{$\theta$}}) &=& \int_{0}^{t} \alpha (\mbox{\boldmath{$v$}}^{{\rm [IJ]}}(t'|\mbox{\boldmath{$\theta$}}) + \mbox{\boldmath{$v$}}^{{\rm [IJ]}}(t'|\mbox{\boldmath{$\theta$}})^{{\rm T}} + \mbox{\boldmath{$c$}}(t')) \ {\rm d} t' \nonumber \\
&=& \int_{0}^{t} \alpha^{2} \int_{0}^{t'} \exp (\mbox{\boldmath{$a$}} (t'-t'')) \ (\mbox{\boldmath{$v$}}^{{\rm [II]}}(t''|\mbox{\boldmath{$\theta$}}) + \mbox{\boldmath{$c$}}(t'')) + \nonumber \\
& & (\mbox{\boldmath{$v$}}^{{\rm [II]}}(t''|\mbox{\boldmath{$\theta$}}) + \mbox{\boldmath{$c$}}(t'')) \ \exp (\mbox{\boldmath{$a$}}^{{\rm T}} (t'-t'')) \ {\rm d} t'' + \alpha \mbox{\boldmath{$c$}}(t') \ {\rm d} t'.
\label{vJJijsol}
\end{eqnarray}

\section{Moments of $I$ and $J$}
\label{moIJ}

Eq.(\ref{mIevo}) through (\ref{vJJevo}) are the differential equations for the time evolution of the first and second order moments of $I$ and $J$. The symbols $m^{{\rm [I]}}(t|\mbox{\boldmath{$\theta$}})$, $m^{{\rm [J]}}(t|\mbox{\boldmath{$\theta$}})$ are the mean of $I$, $J$, and $v^{{\rm [II]}}(t|\mbox{\boldmath{$\theta$}})$, $v^{{\rm [IJ]}}(t|\mbox{\boldmath{$\theta$}})$, $v^{{\rm [JJ]}}(t|\mbox{\boldmath{$\theta$}})$ are the variance of $I$, covariance between $I$ and $J$, variance of $J$.
\begin{eqnarray}
\frac{{\rm d}m^{{\rm [I]}}(t|\mbox{\boldmath{$\theta$}})}{{\rm d}t} = (\alpha-\beta) m^{{\rm [I]}}(t|\mbox{\boldmath{$\theta$}}).
\label{mIevo}
\end{eqnarray}
\begin{eqnarray}
\frac{{\rm d}m^{{\rm [J]}}(t|\mbox{\boldmath{$\theta$}})}{{\rm d}t} = \alpha m^{{\rm [I]}}(t|\mbox{\boldmath{$\theta$}}).
\label{mJevo}
\end{eqnarray}
\begin{eqnarray}
\frac{{\rm d} v^{{\rm [II]}}(t|\mbox{\boldmath{$\theta$}})}{{\rm d} t} = 2(\alpha-\beta) v^{{\rm [II]}}(t|\mbox{\boldmath{$\theta$}}) + (\alpha+\beta) m^{{\rm [I]}}(t|\mbox{\boldmath{$\theta$}}).
\label{vIIevo}
\end{eqnarray}
\begin{eqnarray}
\frac{{\rm d} v^{{\rm [IJ]}}(t|\mbox{\boldmath{$\theta$}})}{{\rm d} t} = (\alpha-\beta) v^{{\rm [IJ]}}(t|\mbox{\boldmath{$\theta$}}) + \alpha (v^{{\rm [II]}}(t|\mbox{\boldmath{$\theta$}}) + m^{{\rm [I]}}(t|\mbox{\boldmath{$\theta$}})).
\label{vIJevo}
\end{eqnarray}
\begin{eqnarray}
\frac{{\rm d} v^{{\rm [JJ]}}(t|\mbox{\boldmath{$\theta$}})}{{\rm d} t} = \alpha (2v^{{\rm [IJ]}}(t|\mbox{\boldmath{$\theta$}}) + m^{{\rm [I]}}(t|\mbox{\boldmath{$\theta$}})).
\label{vJJevo}
\end{eqnarray}

Eq.(\ref{mIsol}) through (\ref{vJJsol}) are the solutions.
\begin{eqnarray}
m^{{\rm [I]}}(t|\mbox{\boldmath{$\theta$}}) =I(0) \exp(\alpha-\beta)t.
\label{mIsol}
\end{eqnarray}
\begin{eqnarray}
m^{{\rm [J]}}(t|\mbox{\boldmath{$\theta$}}) =I(0) (\frac{\alpha}{\alpha-\beta} \exp(\alpha-\beta)t - \frac{\beta}{\alpha-\beta}).
\label{mJsol} 
\end{eqnarray}
\begin{eqnarray}
v^{{\rm [II]}}(t|\mbox{\boldmath{$\theta$}}) =I(0) \frac{\alpha+\beta}{\alpha-\beta} (\exp 2(\alpha-\beta)t - \exp (\alpha-\beta)t).
\label{vIIsol}
\end{eqnarray}
\begin{eqnarray}
v^{{\rm [IJ]}}(t|\mbox{\boldmath{$\theta$}}) &=& I(0) \{ \frac{\alpha(\alpha+\beta)}{(\alpha-\beta)^{2}} \exp 2(\alpha-\beta)t \nonumber \\
&-& (\frac{\alpha(\alpha+\beta)}{(\alpha-\beta)^{2}} + \frac{2\alpha \beta}{\alpha-\beta}t) \exp (\alpha-\beta)t \}.
\label{vIJsol}
\end{eqnarray}
\begin{eqnarray}
v^{{\rm [JJ]}}(t|\mbox{\boldmath{$\theta$}}) &=& I(0) [ \frac{\alpha^{2}(\alpha+\beta)}{(\alpha-\beta)^{3}} \exp 2(\alpha-\beta)t \nonumber \\
&-& \{\frac{\alpha(\alpha+\beta)}{(\alpha-\beta)^{2}} + \frac{4\alpha^{2} \beta}{(\alpha-\beta)^{2} }t\} \exp (\alpha-\beta)t - \frac{\alpha\beta(\alpha+\beta)}{(\alpha-\beta)^{3}} ].
\label{vJJsol}
\end{eqnarray}

\end{document}